\documentclass[sigconf,nonacm]{acmart}

\renewcommand\footnotetextcopyrightpermission[1]{}
\setcopyright{none}
\usepackage{booktabs}
\usepackage{tabularx}
\usepackage{array}
\usepackage{amsmath}
\usepackage{microtype}

\acmConference[Converted Manuscript]{Converted Manuscript}{2023}{Beijing, China}
\acmYear{2023}
\acmISBN{}
\acmDOI{}

\title{Entity Alignment Method of Science and Technology Patent Based on Graph Convolution Network and Information Fusion}

\author{Runze Fang}
\affiliation{%
  \institution{Beijing Key Laboratory of Intelligent Communication Software and Multimedia, School of Computer Science (National Pilot Software Engineering School), Beijing University of Posts and Telecommunications}
  \city{Beijing}
  \country{China}
}

\author{Yawen Li}
\authornote{Corresponding author.}
\email{warmly0716@126.com}
\affiliation{%
  \institution{School of Economics and Management, Beijing University of Posts and Telecommunications}
  \city{Beijing}
  \country{China}
}

\author{Yingxia Shao}
\affiliation{%
  \institution{Beijing Key Laboratory of Intelligent Communication Software and Multimedia, School of Computer Science (National Pilot Software Engineering School), Beijing University of Posts and Telecommunications}
  \city{Beijing}
  \country{China}
}

\author{Zeli Guan}
\affiliation{%
  \institution{Beijing Key Laboratory of Intelligent Communication Software and Multimedia, School of Computer Science (National Pilot Software Engineering School), Beijing University of Posts and Telecommunications}
  \city{Beijing}
  \country{China}
}

\author{Zhe Xue}
\affiliation{%
  \institution{Beijing Key Laboratory of Intelligent Communication Software and Multimedia, School of Computer Science (National Pilot Software Engineering School), Beijing University of Posts and Telecommunications}
  \city{Beijing}
  \country{China}
}

\begin{document}

\begin{abstract}
The entity alignment of science and technology patents aims to link the equivalent entities in the knowledge graph of different science and technology patent data sources. Most entity alignment methods only use graph neural network to obtain the embedding of graph structure or use attribute text description to obtain semantic representation, ignoring the process of multi-information fusion in science and technology patents. In order to make use of the graphic structure and auxiliary information such as the name, description and attribute of the patent entity, this paper proposes an entity alignment method based on the graph convolution network for science and technology patent information fusion. Through the graph convolution network and BERT model, the structure information and entity attribute information of the science and technology patent knowledge graph are embedded and represented to achieve multi-information fusion, thus improving the performance of entity alignment. Experiments on three benchmark data sets show that the proposed method has better Hits@$K$ evaluation indicators than existing methods.
\end{abstract}


\maketitle

\section{Introduction}

With the rapid development of science and technology and knowledge map research \cite{huang2019survey}, a large number of science and technology patents have emerged. Broader innovation studies also indicate that scientific and technological resources interact with education, identity and creative capacity in innovation processes \cite{zhou2020creativeEntrepreneur}. Scientific resources are often multi-view and dynamic: scholar clustering work has shown that research interests can evolve across views \cite{li2023multiViewScholar}, and scientific information retrieval must handle both semantic and media heterogeneity \cite{li2022crossMediaRetrieval}. However, at present, many knowledge graphs \cite{li2021heterogeneousLatentTopic} related to science and technology patents are constructed by different institutions and individuals. Large graph processing also faces memory and scalability constraints, as shown by second-order random walk research on billion-edge natural graphs \cite{shao2021memoryAwareRandomWalk}. The requirements of these knowledge graphs are specific, and the design and construction are not uniform \cite{meng2020overview}, so there are problems of heterogeneity and redundancy among them. The entity alignment of science and technology patents is the key technology in the process of knowledge fusion of science and technology patents. The main purpose is to find the equivalent entities between different science and technology patent data. Because the knowledge content of different science and technology patent data has different sources and human understanding, the text expression of the same thing in different data sources will be different, so there are significant problems in the integration of science and technology patent knowledge graph.

With the development of entity alignment technology \cite{trisedya2019attributeEmbeddings}, many scholars have proposed different kinds of entity alignment methods for science and technology patents, and a large number of entity alignment research documents have emerged \cite{kou2018hashtag}. Heterogeneous graph attention networks further show that typed nodes and relations can improve representation learning when supervision is limited \cite{hu2019heterogeneousGraphAttention}. In recent years, with the rapid development of knowledge representation learning technology, researchers have proposed a number of knowledge representation technology patent entity alignment methods based on deep learning \cite{zhao2020experimental}. This method uses knowledge representation learning technology such as graph neural network or graph convolution network to embed the knowledge graph, and maps different embedding spaces to the same vector space according to the aligned entities \cite{guan2021federatedGraph}. Some of these methods integrate other elements of scientific and technological patent entities into vector space through multi-information fusion \cite{gao2021mhgcn}, and get the result of entity alignment according to the distance or similarity of entities in vector space.

Because graph convolution network can capture the attribute characteristics and adjacency information between scientific and technological patent entities, and has strong robustness and generalization ability, the entity alignment method based on graph convolution network has a good effect. However, the existing method of entity alignment of scientific and technological patent documents \cite{cheng2022multijaf} does not fully use the various information provided by scientific and technological patents, and the method of entity alignment of scientific and technological patents has a lot of room for improvement. Federated self-adaptive learning for information network representation also suggests that decentralized scientific resources can be modeled without forcing all data into a single platform \cite{li2026fedSIN}, while federated learning with stochastic quantization provides a related direction for reducing communication and computation costs in distributed learning settings \cite{li2022stochasticQuantization}. Therefore, the integration of entities from different sources of scientific and technological patents is an urgent problem to be solved.

The existing entity alignment methods for science and technology patents \cite{gao2021hyperea} still have shortcomings in multi-information fusion. Most of them \cite{chen2020incidental} are only based on deep learning for structure embedding or based on word bag model for attribute embedding, and few of them fuse the embedded information of the two. Graph neural networks for graphs with incomplete features and structures indicate that missing or noisy information should be handled jointly rather than ignored \cite{huo2023t2gnn}. Few multi-information fusion methods \cite{meng2013tracking} also mostly use the attribute description representation method with poor effect.

In this paper, we propose an entity alignment method for science and technology patents based on graph convolution network and information fusion. The graph convolution network is used to deeply mine the structure information of knowledge graph, including topological connection, relationship and other information. In addition to the structural information, the entity attribute information in the scientific and technological patent is represented in depth semantics based on BERT to improve the effect of entity alignment.

The main contributions of this paper include three aspects:
\begin{enumerate}
  \item A new entity alignment method for science and technology patent information fusion based on graph convolution network is proposed.
  \item Based on the multi-level alignment method, the topological structure, relationship and attribute information are integrated into the graph convolution network.
  \item A multi-information fusion method of science and technology patents based on graph structure embedding and attribute semantic representation is presented.
\end{enumerate}

\section{Related Work}

Most of the traditional entity alignment methods of science and technology patents focus on syntax and structure \cite{yan2020ctea}, especially the early entity alignment and mapping technology of science and technology patents mainly focus on calculating the distance between labels and characters \cite{lin2009average}. The traditional entity alignment method \cite{li2022vehicleFuel} of science and technology patents mainly solves the entity alignment problem from two aspects: one is based on similarity calculation to compare the symbolic features of entities, and the other is based on relational reasoning. Related interpretable machine learning studies for intelligent decision-making also show that feature semantics should remain understandable when models are used in practical decision scenarios \cite{li2019interpretableDecision}. Recent research also uses statistical machine learning to improve accuracy. However, due to the different attributes of entities and the different fields involved \cite{chen2020seeds} in the process of entity alignment of science and technology patents, it is difficult to give a unified similarity calculation function. At the same time, this discrete attribute information ignores the semantic information \cite{li2013gaussianPHD} implied in many aspects, making the alignment effect limited. Therefore, in recent years, more and more scholars have begun to propose a new technology of entity alignment of science and technology patent based on deep learning.

The entity alignment method of science and technology patent based on graph neural network \cite{yuan2020financialSentiment} has become the main technology to solve the entity alignment problem of science and technology patent, and has achieved good results. Most of these methods \cite{peng2020embedding} use translation model or graph convolution network for knowledge representation learning, because they have strong robustness and generalization ability. The literature uses graph convolution network to embed the entities and relationships of science and technology patents. This model \cite{tang2020bertInt} combines the advantages of graph convolution network and the translation property of the knowledge graph of science and technology patents. It uses convolution function \cite{yang2019aligning} to distinguish the entities of science and technology patents with different roles and effectively capture the structural information, but this method ignores the auxiliary role of the attribute description information of science and technology patents. Semantic-similarity attention with hypergraph convolution has also been explored for scientific publication representation learning, which is closely related to capturing higher-order scientific relations \cite{li2026semanticSimilarityHypergraph}.

The NMN (Neighborhood Matching Network) model \cite{guan2019selfLearning} is proposed in the literature. When the graph convolution network is embedded, the cross-graph attention mechanism is used to obtain the differences of adjacent technological patent entities. Federated learning for supervised cross-modal retrieval provides a related fusion setting in which supervision aligns heterogeneous representations under distributed data constraints \cite{li2024federatedCrossModalRetrieval}. Heterogeneous graph neural networks based on self-supervised reciprocally contrastive learning also suggest that contrastive signals can improve representation consistency across graph views \cite{jin2022heterogeneousGraphContrastive}. The literature \cite{liu2021activeEA} proposed that the short distance difference and long distance dependence of adjacent scientific and technological patent entities can be captured using graph convolution network and self-attention mechanism, but this method needs a long time to capture the dependence information, and the operation efficiency of the model is not high.

Obtaining structure information based on graph convolution network will ignore the attribute information of science and technology patent entities \cite{mao2020relational}, and realizing multi-information fusion of structure information and attribute information can improve the effect of science and technology patent entity alignment. Community detection with deep learning demonstrates that modular graph structure can provide useful high-level signals for representation learning \cite{yang2016modularityDeepLearning}. The JAPE algorithm is proposed in the literature. The algorithm \cite{li2020leveragingGraph} includes two modules, namely, the structure embedding based on the relation triplet and the attribute embedding based on the attribute triplet. Self-supervised graph co-training in session-based recommendation further supports using complementary graph views when labels are limited \cite{xia2021selfSupervisedGraph}. The entity alignment of scientific and technological patents \cite{liang2020abstractiveSocial} is achieved through the multi-information fusion of the two modules. However, the method \cite{meng2016consensus} is based on the graph structure embedding of scientific and technological patents based on the TransE model, and the accuracy of entity alignment is not high.

The literature \cite{xu2016multiFeatureIndexing} puts forward the CTEA framework, which is based on the TransE model to generate the structure embedding and context representation of the structure information of the scientific and technological patent entity. According to the attribute information of the scientific and technological patent entity, the candidate set is generated for each source entity and the subject information of the attribute information is captured. Efficient sequence modeling such as filter-enhanced MLPs is less directly related to entity alignment, but it offers useful evidence that lightweight feature filtering can improve representation tasks \cite{zhou2022filterEnhancedMLP}; dataset distillation for sequential recommendation is another related data-efficiency problem, where compact training data are used to preserve model performance \cite{zhang2025td3}. However, this method \cite{yang2015ontology} has not been able to obtain the deep semantic representation of the attribute information. Retrieval-oriented masked autoencoder pre-training also suggests that language models should preserve ranking-sensitive semantics for downstream retrieval and matching tasks \cite{xiao2022retromae}. The document proposes \cite{xu2022topicTrend} the MultiKE (Multi-view KG Embedding) framework, which integrates the structure information, attribute information \cite{li2014phdGlint} and entity name information at the same time. Incomplete multi-view learning work based on view-category interactive sharing transformers also shows that interaction among views and labels is useful when information sources are not complete \cite{ou2024viewCategoryTransformer}. The framework \cite{xu2023infomax} is combined based on the shared space, and each embedding is orthogonally mapped to the same space to achieve multi-information fusion. However, the framework \cite{chen2020stockAttention} is limited to general data sets and is not applicable to the field of scientific and technological patent documents \cite{cui2021mvgan}.

\section{Entity Alignment Method Based on Graph Convolution Network and Information Fusion}

\subsection{MIFEA Method Proposed}

This paper proposes an entity alignment method for science and technology patents based on graph convolution network and information fusion (MIFEA). By integrating the graph structure embedding of science and technology patents and the deep semantic information of the entity attributes of science and technology patents, mapping to the same vector space for entity alignment improves the accuracy of entity alignment for science and technology patents. The method is mainly composed of four parts, including the embedded representation of the structure of the technology patent diagram, the semantic representation of the technology patent attribute, the multi-information fusion and the alignment of the technology patent entity.

The graph structure embedding representation module uses the graph convolution network to deeply mine the structure information of the knowledge graph, including topology connection, relationship and other information. The attribute semantic representation module performs deep semantic representation of the name and description information of entity attributes. The multi-information fusion module combines the graph structure embedding and attribute semantic representation to obtain the vector representation of the scientific and technological patent entity. The entity alignment module obtains the set of similar entities of each technology patent entity through similarity calculation and distance ranking.

\subsection{Embedded Representation of Structure of Science and Technology Patent Map}

MIFEA captures information from the graph structure of the graph of scientific and technological patent knowledge based on graph convolution network, and multi-layer graph convolution stack collects features from multi-hop neighbors. GCN can iteratively update the representation of each entity node through the propagation mechanism in the graph. The equivalent entities of science and technology patents tend to be adjacent to the equivalent entities through the same type of relationship, and the equivalent entities of science and technology patents tend to share similar or even the same attributes. GCN allows end-to-end learning of predictive pipes with arbitrary size and shape input. The input of GCN is the eigenvector of nodes and the structure of graph; the goal of GCN is to learn the characteristic function on the input graph and generate node-level output.

The graph convolution network part of this paper is composed of several stacked GCN layers. The input of layer $l$ of the GCN model is the vertex characteristic matrix $H^{(l)} \in R^{n \times d(l)}$, where $n$ is the number of vertices and $d(l)$ is the number of features in layer $l$. Through the following convolution calculation, the output of layer $l$ is the new characteristic matrix $H^{(l)}$, as shown in equation (1):
\begin{equation}
H^{(l+1)} = \sigma(\hat{D}^{-\frac{1}{2}}\hat{A}\hat{D}^{-\frac{1}{2}}H^{(l)}W^{(l)}).
\end{equation}
Where $\sigma$ is the activation function, $A$ is the connectivity matrix representing the structural information of the science and technology patent graph, $D$ is the diagonal node degree matrix of $A$, $W$ is the weight matrix of the $l$ layer of GCN, and $d(l+1)$ is the dimension of the new fixed point feature.

MIFEA uses two 2-tier GCNs, each of which processes a graph of scientific and technological patent knowledge to generate the structural embedding of its scientific and technological patent entities. For example, two KGs are represented as $G_1=(E_1,R_1,T_1)$ and $G_2=(E_2,R_2,T_2)$; the corresponding GCN models are expressed as $GCN_1$ and $GCN_2$. For the structural feature vector of scientific and technological patent entities, set the dimension of the feature vector to $d$ in all layers of $GCN_1$ and $GCN_2$. The two GCN models share the weight matrices $W_s^{(1)}$ and $W_s^{(2)}$ of the two-layer structure features. The final output of the two GCNs is the $d$-dimension embedding of the science and technology patent entities, which is further used to find the alignment of the science and technology patent entities.

In order to align a pair of science and technology patent entities, if the relationship and attributes of adjacent science and technology patent entities are used, noise may be introduced. It is a better choice to focus only on the relationship and attribute characteristics of current scientific and technological patent entities. Therefore, this paper proposes a hybrid multidimensional alignment network to better simulate these different features. The topological embedding $H^{(l)}$ is obtained by using the $l$ layer, and then the embedding of relations and attributes is obtained by using the feedforward neural network. The feedforward neural network consists of a full connection layer and a high-speed gate layer. The reason why the algorithm uses the high-speed gate layer is that the high-speed gate layer is generally better than the full-connection layer in terms of convergence speed and effectiveness. The feedforward neural network is defined as formula (2), (3) and (4):
\begin{align}
S_f &= \phi(W_f^{(1)}X_f + b_f^{(1)}),\\
T_f &= \sigma(W_f^{t}S_f + b_f^{t}),\\
G_f &= \phi(W_f^{(2)}S_f + b_f^{(2)})\cdot T_f + S_f\cdot(1-T_f).
\end{align}
Where $f \in \{r,a\}$ and $X_f$ represent a certain aspect (i.e. relationship or attribute) and its original characteristics, $W_f^{(1,2,t)}$ and $b_f^{(1,2,t)}$ are model parameters, $\phi(\cdot)$ is ReLU function, and $\sigma(\cdot)$ is sigmoid function. Based on this, the hybrid multi-aspect entity embedding $H_y=[H_t^{(l)}\oplus G_r\oplus G_a]$ is obtained and further normalized.

Given two technical patent knowledge graphs $G_1=(E_1,R_1,T_1)$ and $G_2=(E_2,R_2,T_2)$ and a group of pre-aligned entity $I(G_1,G_2)$ pairs as training data, the algorithm is trained in a supervised manner. In the training phase, the goal is to embed the cross-language technology patent entities into the same low-dimensional vector space, where the equivalent technology patent entities are close to each other. The loss function is defined as equation (5):
\begin{equation}
Loss=\sum_{(e_1,e_2)\in I}\sum_{(e'_1,e'_2)\in I'}[\rho(h_{e_1},h_{e_2})+\beta-\rho(h_{e'_1},h_{e'_2})].
\end{equation}

\subsection{Attribute Semantic Representation of Science and Technology Patent Entities}

Because different science and technology patent knowledge maps provide text descriptions of entities expressed in different languages, and contain detailed semantic information about entities. The key of attribute semantic representation module is to judge whether the literal description of the equivalent entity of scientific and technological patent is semantically close. However, it is very difficult to directly measure the semantic relevance of the descriptions of two science and technology patent entities, because they are expressed in different languages. Based on the BERT model, MIFEA maps words or sentences in different languages into the same semantic space to bridge the gap between different language descriptions.

MIFEA follows the basic design of BERT, and converts the task of science and technology patent entity to its text matching task. For the two entities $e_1$ and $e_2$ of two science and technology patent knowledge graphs from different data sources, the descriptions in this paper are $t_1$ and $t_2$, which are respectively composed of word sequences in different languages. The attribute semantic representation module consists of two components, each of which takes the description of an entity (from the source language or target language) as input. The input is designed as the format of BERT input, and then fed to the attribute semantic representation for context encoding.

By constructing training data $D=\{(e,e'^{+},e'^{-})\}$, each triplet $(e,e'^{+},e'^{-})\in D$ contains a query technology patent entity $e\in E$, a correctly aligned entity $e'^{+}\in E'$ and a negative aligned entity $e'^{-}\in E'$ randomly selected from $E'$. For each science and technology patent entity in the data set, the pre-trained multi-language BERT accepts its name and description as input, and the CLS of BERT that has been filtered by MLP is embedded into the formula (6):
\begin{equation}
C(e)=MLP(CLS(e)).
\end{equation}
The marginal loss of two pairs is used as the loss function of training, as shown in equation (7):
\begin{equation}
L=\sum_{(e,e'^{+},e'^{-})\in D}\max\{0,g(e,e^{-})-g(e,e^{+})+m\}.
\end{equation}

\subsection{Technology Patent Information Fusion}

This paper uses multiple information fusion to align scientific and technological patent entities, such as entity labels, topology information, relationship types, attributes and text descriptions. The graph convolution network is used to embed the structure information of scientific and technological patents, while adding various information. The topological information, relationship and attribute information are used as the input of the graph convolution network at the same time, not only as the external auxiliary information. Finally, the text description information of the science and technology patent entity is used for BERT embedding learning, and the structure embedding and text description information are fused to prepare for the subsequent alignment of the science and technology patent entity.

The first two modules of MIFEA extract the embedding representation from the graph structure of the science and technology patent knowledge graph and the attribute text description of the science and technology patent entity, that is, graph and text embedding. The multi-information fusion module will integrate the two kinds of embedding, and then align the scientific and technological patent entities. Use multi-layer GCN to obtain the topology information of science and technology patents, and use the full connection layer and high-speed gate mechanism to embed the entity relationship and attributes of science and technology patents. The technology patent text embedded by BERT is represented as $T^B$, and the technology patent knowledge graph embedded by graph convolution network is represented as $T^G$, which is combined based on the weight mechanism, and the two embedded weights are added as the entity embedding, as shown in formula (8):
\begin{equation}
T^C=\tau\cdot T^G\oplus(1-\tau)\cdot T^B.
\end{equation}

\subsection{Alignment of Science and Technology Patent Entities}

After obtaining the embedding of scientific and technological patent entities, cosine similarity is used to measure the distance between candidate entity pairs. The small distance reflects the high probability of entity alignment as an equivalent entity. All candidates are ranked for evaluation.

MIFEA takes the connection of the two descriptions of each candidate entity pair as the input to process each possible pair in the training set. At the same time, in order to reduce the calculation cost, it ignores the candidate pairs with low alignment probability. Sampling based on uncertainty can provide additional ranking improvement. MIFEA uses the model based on graph convolution network to generate a candidate pool whose size is far smaller than the entire entity pool. The model based on graph convolution network provides each source entity with the first $Q$ candidate objects of the target entity (where $Q$ is a super parameter). MIFEA generates a ranking score for each candidate entity pair in the pool to further rank the candidates.

\section{Experiment of MIFEA}

\subsection{Data Sets and Evaluation Indicators}

This paper uses the cross-language data sets DBP15K and DBP100K as the data sets. This paper also uses the technology patent data sets, including patent data in both Chinese and English, with a total of 160000 entries. DBP15K and DBP100K are large knowledge graphs that extract structured knowledge from Wikipedia in multiple languages. DBP15K and DBP100K contain three subsets: ZH-EN, JA-EN and FR-EN. Each subset contains information such as entity, relationship, attribute, relationship triplet and attribute triplet.

The evaluation index is Hits@$K$. It refers to the proportion of correct entities in the top $k$ of the results. The greater the Hits@$K$, the better the effect. This paper compares MIFEA with MTransE, JAPE, BootEA, TransEdge and AttrGNN.

\subsection{Results and Analysis of Main Experiment}

Based on the general data sets DBP15K, DBP100K and scientific and technological patent data sets, the experimental results and algorithm efficiency of different entity alignment algorithms are compared and analyzed. In this paper, the same segmentation settings are used in the experiment, in which 30\% of the data is used for training and the remaining 70\% is used for evaluation. Hits@$K$ is used as an evaluation index; this method measures the proportion of correctly aligned entities in the top $k$ candidate objects.

\begin{table*}[t]
  \caption{Comparison of entity alignment effect}
  \label{tab:entity-alignment-1}
  \begin{tabularx}{\textwidth}{@{}l *{4}{>{\centering\arraybackslash}X}@{}}
    \toprule
    & \multicolumn{2}{c}{DBP15K} & \multicolumn{2}{c}{Science and technology patent} \\
    \cmidrule(lr){2-3}\cmidrule(l){4-5}
    Method & Hits@1 & Hits@10 & Hits@1 & Hits@10 \\
    \midrule
    MTransE & 30.83 & 61.41 & 29.35 & 57.19 \\
    JAPE & 41.18 & 74.46 & 38.46 & 71.31 \\
    BootEA & 47.77 & 83.50 & 44.78 & 80.65 \\
    TransEdge & 73.51 & 91.92 & 71.61 & 89.26 \\
    AttGNN & 79.60 & 92.93 & 78.09 & 90.82 \\
    MIFEA & 82.94 & 94.12 & 80.74 & 92.59 \\
    \bottomrule
  \end{tabularx}
\end{table*}

From Table~\ref{tab:entity-alignment-1} and Table~\ref{tab:entity-alignment-2}, it can be seen that the MTransE model has the worst effect. From Table~\ref{tab:entity-alignment-2}, we can see that because MTransE only uses the structural information of the knowledge graph and uses TransE to construct a relational triplet, it lacks consideration of the overall structural information of the knowledge graph, thus losing some structural information, so the alignment effect of MTransE experiment is poor. Based on MTransE, JAPE integrates attribute information for entity alignment. Hits@1 and Hits@10 are 47\% and 11\% higher than MTransE, respectively. BootEA adds bootstrap training strategy on the basis of TransE, and its effect is also slightly higher than MTransE model. TransEdge is a good model based on the TransE embedding method. The main reason is that TransEdge not only embeds the vector of the head and tail entities of the relationship triple, but also adds the relationship embedding between entities to better represent the knowledge graph. Hits@1 and Hits@10 are 57\% and 11\% higher than BootEA, respectively.

\begin{table*}[t]
  \caption{Comparison of entity alignment effect}
  \label{tab:entity-alignment-2}
  \begin{tabularx}{\textwidth}{@{}l *{4}{>{\centering\arraybackslash}X}@{}}
    \toprule
    & \multicolumn{2}{c}{ZH-EN} & \multicolumn{2}{c}{FR-EN} \\
    \cmidrule(lr){2-3}\cmidrule(l){4-5}
    Method & Hits@1 & Hits@10 & Hits@1 & Hits@10 \\
    \midrule
    MTransE & 26.12 & 57.64 & 21.17 & 53.64 \\
    JAPE & 38.52 & 64.32 & 27.61 & 54.44 \\
    BootEA & 44.04 & 78.58 & 61.82 & 84.17 \\
    TransEdge & 69.53 & 87.26 & 67.59 & 91.32 \\
    AttGNN & 77.65 & 89.37 & 84.69 & 93.45 \\
    MIFEA & 79.90 & 91.34 & 89.73 & 94.18 \\
    \bottomrule
  \end{tabularx}
\end{table*}

It can be seen from Table~\ref{tab:entity-alignment-2} that AttGNN performs well in the experiment and reaches the best in the FR-EN dataset Hits@10. This is 1.3\% higher than the MIFEA proposed in this paper. The main reason is that AttGNN is a model based on the GNN embedding method. This embedding method recursively aggregates the eigenvectors of neighbor nodes, which can take account of attribute information and structure information, and make good use of the data information in the knowledge graph.

The algorithm MIFEA proposed in this paper is optimal on ZH-EN and FR-EN data sets Hits@1 and Hits@10. Compared with AttGNN, ZH-EN increased by 2.8\% and 2.2\% respectively. The main reason is that MIFEA uses the description information of entities, and uses BERT to train the description information. Through the description information, the semantic information of entities can be well obtained, so as to better distinguish. For example, the two language syntax of FR-EN dataset is similar, so MIFEA performs best in this dataset.

\subsection{Results and Analysis of MIFEA Ablation Experiment}

This paper carries out ablation experiments on the proposed algorithm to verify the effectiveness of each technology patent knowledge graph information component. Select to remove the features of different aspects of the structure semantic representation module (SE), namely relations, attributes and high-speed gate mechanism. These variants are represented as $SE_{w/oRE}$, $SE_{w/oAE}$ and $SE_{w/oHW}$ respectively.

\begin{table*}[t]
  \caption{Ablation experiment of MIFEA}
  \label{tab:ablation}
  \begin{tabularx}{\textwidth}{@{}l *{4}{>{\centering\arraybackslash}X}@{}}
    \toprule
    & \multicolumn{2}{c}{ZH-EN} & \multicolumn{2}{c}{JA-EN} \\
    \cmidrule(lr){2-3}\cmidrule(l){4-5}
    Method & Hits@1 & Hits@10 & Hits@1 & Hits@10 \\
    \midrule
    $SE_{w/oRE}$ & 50.2 & 78.4 & 52.6 & 81.6 \\
    $SE_{w/oAE}$ & 49.2 & 81.0 & 52.2 & 83.3 \\
    $SE_{w/oHW}$ & 46.8 & 76.1 & 50.5 & 79.5 \\
    SE & 56.2 & 85.1 & 56.7 & 86.9 \\
    \bottomrule
  \end{tabularx}
\end{table*}

The experimental results are shown in Table~\ref{tab:ablation}. It can be seen that the performance of the data set is reduced after deleting the relationship or attribute feature. Therefore, when MIFEA deletes topological features, the algorithm loses this structural knowledge, so its results are worse. From these experiments, we can conclude that topology information plays an indispensable role in making alignment decisions. When the high-speed gate mechanism is deleted, the effect of the algorithm decreases the most, which shows that the high-speed gate mechanism plays an important role in the convergence speed and effectiveness of the neural model used for entity matching.

\subsection{Operation Efficiency Analysis}

In order to compare and analyze the efficiency of MIFEA methods, this paper will conduct experiments on different algorithms on a unified platform based on scientific and technological patent data sets. Each algorithm is run five times on the ZH-EN data set based on the scientific and technological patent data set to obtain the average running time. The efficiency of different algorithms is compared and analyzed through the running results.

\begin{figure}[t]
  \centering
  \includegraphics[width=\linewidth]{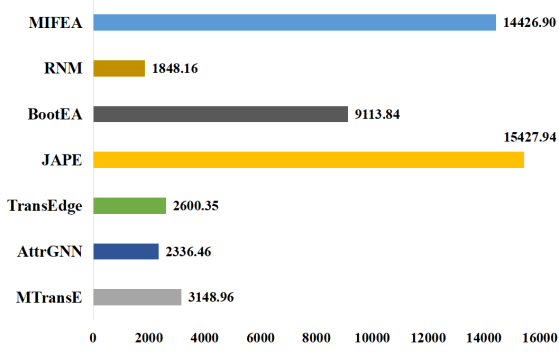}
  \caption{Algorithm running time diagram}
  \Description{Bar chart comparing the running time of MIFEA, RNM, BootEA, JAPE, TransEdge, AttrGNN and MTransE.}
  \label{fig:running-time}
\end{figure}

The result of algorithm running time is shown in Figure~\ref{fig:running-time}. From the figure, we can see that the RNM algorithm is efficient, and its running time is only 11.9\% of the JAPE running time. It achieves good results with less space and less time. This shows that the subgraph matching based on relationship and the probability calculation of adjacent entities are very suitable for the entity alignment task of knowledge graph. The running time of JAPE algorithm is the longest, which indicates that the efficiency of the embedding method based on TransE is not high, and the fusion of attribute information also increases the complexity of the algorithm, thus increasing the running time.

The algorithm MIFEA proposed in this paper also has a long running time, but as an information fusion algorithm, the efficiency of MIFEA is 6.4\% higher than that of JAPE. The algorithm uses BERT's model to obtain the deep semantic representation of attribute description information. Although it can effectively improve the entity alignment effect, the universality and fine-tuning time of the model need to be improved.

\subsection{Proportion Analysis of Training Set}

In order to study how the size of the training set affects the results of the algorithm, different numbers of pre-aligned entities are used as training data to further compare MIFEA with JAPE, TransEdge and AttGNN. The experimental results show that the more pre-aligned entities used, the better the results of the algorithm.

\begin{figure}[t]
  \centering
  \includegraphics[width=\linewidth]{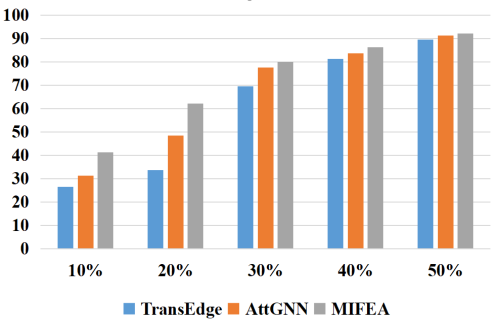}\\[-0.4em]
  {\small (a) ZH-EN}\par\medskip
  \includegraphics[width=\linewidth]{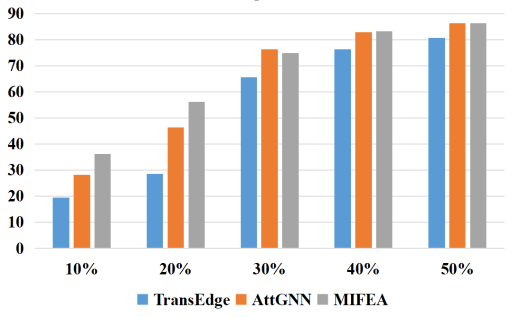}\\[-0.4em]
  {\small (b) JA-EN}
  \caption{Experimental results of different training sets}
  \Description{Two bar charts showing entity alignment performance under different proportions of training data on ZH-EN and JA-EN.}
  \label{fig:training-set}
\end{figure}

Pre-aligned entities of different proportions are used as training data, with a range of 10\% to 50\%. All remaining pre-aligned entities are used for testing. Figure~\ref{fig:training-set} shows the results of three methods in two data sets Hits@1. The results show that with the increase of training data, these three methods perform better. In the ZH-EN dataset, MIFEA is always better than the other two algorithms. The higher the proportion of training data, the smaller the gap between the results of the algorithms. In JA-EN dataset, MIFEA is always better than TransEdge, and the result is close to AttGNN. Due to some differences in the semantics of Japanese and Chinese, the BERT pre-training model adopted by MIFEA has not fully played its role, and the embedded representation of attribute semantics has not shown advantages, so the results are not better than those of AttGNN.

\section{Conclusion}

This paper proposes a method of entity alignment of science and technology patents based on graph convolution network and information fusion, so as to find the mapping of equivalent entities in the knowledge graph of science and technology patents. A method based on graph convolution network and BERT is proposed to study how to make better use of various information of entities provided in scientific and technological patent data, including topological connection, relationship, attribute, entity label and text description. The graph convolution network is used to embed the structure information while fusing various information. A multi-level alignment method is proposed, which uses topology information, relationship and attribute information as the input of the graph convolution network at the same time. The text description information is used to realize the deep semantic representation by BERT, and the entity alignment is achieved by integrating the graph structure embedded representation and the text semantic representation. The experimental results show that the method proposed in this paper achieves the best performance on the scientific and technological patent data set. With the increasing complexity and scale of scientific and technological patent data sources, it is necessary to consider the implementation efficiency and accuracy of the method.

\begin{acks}
This work was supported by the National Natural Science Foundation of China (62172056, U22B2038, 62192784).
\end{acks}

\bibliographystyle{gbt7714-numerical}
\bibliography{references}

@article{li2023multiViewScholar,
  author = {Ang Li and Yawen Li and Yingxia Shao and Bingyan Liu},
  title = {Multi-View Scholar Clustering With Dynamic Interest Tracking},
  journal = {{IEEE} Transactions on Knowledge and Data Engineering},
  year = {2023},
  volume = {35},
  number = {9},
  pages = {9671--9684},
  language = {english}
}

@article{li2021heterogeneousLatentTopic,
  author = {Yawen Li and Di Jiang and Rongzhong Lian and Xueyang Wu and Conghui Tan and Yi Xu and Zhiyang Su},
  title = {Heterogeneous Latent Topic Discovery for Semantic Text Mining},
  journal = {{IEEE} Transactions on Knowledge and Data Engineering},
  year = {2021}
}

@article{meng2020overview,
  author = {Pengbo Meng},
  title = {Overview of Entity Alignment Based on Graph Neural Network},
  journal = {Modern Computer},
  year = {2020},
  number = {9},
  pages = {37--40}
}

@article{huang2019survey,
  author = {Hengqi Huang and Juan Yu},
  title = {A Survey of Knowledge Map Research},
  journal = {Computer System Application},
  year = {2019},
  volume = {28},
  number = {6},
  pages = {1--12}
}

@article{li2022crossMediaRetrieval,
  author = {Ang Li and Junping Du and Feifei Kou and Zhe Xue and Xin Xu and Mingying Xu and Yang Jiang},
  title = {Scientific and Technological Information Oriented Semantics-Adversarial and Media-Adversarial Cross-Media Retrieval},
  journal = {arXiv preprint arXiv:2203.08615},
  year = {2022},
  language = {english}
}

@article{li2026fedSIN,
  author = {Ang Li and Yawen Li and Zhe Xue},
  title = {{FedSIN}: Information Network Representation Based on Federated Self-Adaptive Learning},
  journal = {Frontiers of Computer Science},
  year = {2026},
  volume = {20},
  number = {1},
  pages = {2001307},
  language = {english}
}

@article{li2024federatedCrossModalRetrieval,
  author = {Ang Li and Yawen Li and Yingxia Shao},
  title = {Federated Learning for Supervised Cross-Modal Retrieval},
  journal = {World Wide Web},
  year = {2024},
  volume = {27},
  number = {4},
  pages = {41},
  language = {english}
}

@article{li2026semanticSimilarityHypergraph,
  author = {Ang Li and Yawen Li and Feifei Kou and Zhe Xue and Meiyu Liang and Baoxiang Wang},
  title = {Semantic-Similarity Attention Meets Hypergraph Convolution for Scientific Publication Representation Learning},
  journal = {Frontiers of Computer Science},
  year = {2026},
  language = {english}
}

@inproceedings{hu2019heterogeneousGraphAttention,
  author = {Linmei Hu and Tianchi Yang and Chuan Shi and Houye Ji and Xiaoli Li},
  title = {Heterogeneous Graph Attention Networks for Semi-Supervised Short Text Classification},
  booktitle = {Conference on Empirical Methods in Natural Language Processing},
  year = {2019},
  pages = {4821--4830},
  language = {english}
}

@inproceedings{zhou2022filterEnhancedMLP,
  author = {Kun Zhou and Hui Yu and Wayne Xin Zhao and Jirong Wen},
  title = {Filter-Enhanced {MLP} Is All You Need for Sequential Recommendation},
  booktitle = {{ACM} Web Conference},
  year = {2022},
  pages = {2388--2399},
  language = {english}
}

@inproceedings{yang2016modularityDeepLearning,
  author = {Liang Yang and Xiaochun Cao and Dongxiao He and Chuan Wang and Xiao Wang and Weixiong Zhang},
  title = {Modularity Based Community Detection With Deep Learning},
  booktitle = {International Joint Conference on Artificial Intelligence},
  year = {2016},
  pages = {2252--2258},
  language = {english}
}

@inproceedings{xia2021selfSupervisedGraph,
  author = {Xin Xia and Hongzhi Yin and Junliang Yu and Yingxia Shao and Lizhen Cui},
  title = {Self-Supervised Graph Co-Training for Session-Based Recommendation},
  booktitle = {The 30th {ACM} International Conference on Information and Knowledge Management},
  year = {2021},
  pages = {2180--2190},
  language = {english}
}

@inproceedings{huo2023t2gnn,
  author = {Cuiying Huo and Di Jin and Yawen Li and Dongxiao He and Yubin Yang and Lingfei Wu},
  title = {{T2-GNN}: Graph Neural Networks for Graphs With Incomplete Features and Structure via Teacher-Student Distillation},
  booktitle = {Conference on Artificial Intelligence},
  year = {2023},
  pages = {4339--4346},
  language = {english}
}

@article{li2019interpretableDecision,
  author = {Yawen Li and Liu Yang and Bohan Yang and Ning Wang and Tian Wu},
  title = {Application of Interpretable Machine Learning Models for the Intelligent Decision},
  journal = {Neurocomputing},
  year = {2019},
  volume = {333},
  pages = {273--283},
  language = {english}
}

@inproceedings{xiao2022retromae,
  author = {Shitao Xiao and Zheng Liu and Yingxia Shao and Zhao Cao},
  title = {{RetroMAE}: Pre-Training Retrieval-Oriented Language Models via Masked Auto-Encoder},
  booktitle = {Conference on Empirical Methods in Natural Language Processing},
  year = {2022},
  pages = {538--548},
  language = {english}
}

@inproceedings{trisedya2019attributeEmbeddings,
  author = {Bayu Distiawan Trisedya and Jianzhong Qi and Rui Zhang},
  title = {Entity Alignment between Knowledge Graphs Using Attribute Embeddings},
  booktitle = {Proceedings of the {AAAI} Conference on Artificial Intelligence},
  year = {2019},
  volume = {33},
  number = {01},
  pages = {297--304}
}

@article{kou2018hashtag,
  author = {Feifei Kou and Junping Du and Congxian Yang and Yansong Shi and Wanqiu Cui and Meiyu Liang and Yue Geng},
  title = {Hashtag Recommendation Based on Multi-Features of Microblogs},
  journal = {Journal of Computer Science and Technology},
  year = {2018},
  volume = {33},
  pages = {711--726}
}

@article{zhao2020experimental,
  author = {Xi Zhao and Weiqin Zeng and Jiuyang Tang and Wei Wang and Fabian Suchanek},
  title = {An Experimental Study of State-of-the-Art Entity Alignment Approaches},
  journal = {{IEEE} Transactions on Knowledge and Data Engineering},
  year = {2020},
  volume = {34},
  number = {6},
  pages = {2610--2625}
}

@inproceedings{guan2021federatedGraph,
  author = {Zeli Guan and Yawen Li and Zhe Xue and Yuxin Liu and Hongrui Gao and Yingxia Shao},
  title = {Federated Graph Neural Network for Cross-Graph Node Classification},
  booktitle = {{IEEE} International Conference on Cloud Computing and Intelligence Systems},
  year = {2021},
  pages = {418--422},
  language = {english}
}

@article{gao2021mhgcn,
  author = {Jie Gao and Xiaohan Liu and Yuting Chen and others},
  title = {{MHGCN}: Multiview Highway Graph Convolutional Network for Cross-Lingual Entity Alignment},
  journal = {Tsinghua Science and Technology},
  year = {2021},
  volume = {27},
  number = {4},
  pages = {719--728}
}

@article{cheng2022multijaf,
  author = {Bo Cheng and Jing Zhu and Maoguo Gong},
  title = {{MultiJAF}: Multi-Modal Joint Entity Alignment Framework for Multi-Modal Knowledge Graph},
  journal = {Neurocomputing},
  year = {2022},
  volume = {500},
  pages = {581--591}
}

@inproceedings{gao2021hyperea,
  author = {Sheng Gao},
  title = {{HyperEA}: Hyperbolic Entity Alignment between Knowledge Graphs},
  booktitle = {2021 4th International Conference on Artificial Intelligence and Big Data},
  year = {2021},
  pages = {550--554}
}

@article{chen2020incidental,
  author = {Muhao Chen and Weijia Shi and Ben Zhou and Dan Roth},
  title = {Cross-Lingual Entity Alignment with Incidental Supervision},
  journal = {arXiv preprint arXiv:2005.00171},
  year = {2020}
}

@article{meng2013tracking,
  author = {Deyuan Meng and Yingmin Jia and Junping Du and Fashan Yu},
  title = {Tracking Algorithms for Multiagent Systems},
  journal = {{IEEE} Transactions on Neural Networks and Learning Systems},
  year = {2013},
  volume = {24},
  number = {10},
  pages = {1660--1676}
}

@article{yan2020ctea,
  author = {Zhaohui Yan and Rong Peng and Yunqi Wang and others},
  title = {{CTEA}: Context and Topic Enhanced Entity Alignment for Knowledge Graphs},
  journal = {Neurocomputing},
  year = {2020},
  volume = {410},
  pages = {419--431}
}

@inproceedings{lin2009average,
  author = {Peng Lin and Yingmin Jia and Junping Du and Fashan Yu},
  title = {Average Consensus for Networks of Continuous-Time Agents with Delayed Information and Jointly-Connected Topologies},
  booktitle = {2009 American Control Conference},
  year = {2009},
  pages = {3884--3889}
}

@article{li2022vehicleFuel,
  author = {Yawen Li and Isabella Yunfei Zeng and Ziheng Niu and Jiahao Shi and Ziyang Wang and Zeli Guan},
  title = {Predicting Vehicle Fuel Consumption Based on Multi-View Deep Neural Network},
  journal = {Neurocomputing},
  year = {2022},
  volume = {502},
  pages = {140--147}
}

@inproceedings{chen2020seeds,
  author = {Xiaojun Chen and Limin Wang and Yanchun Tang and others},
  title = {Seeds Optimization for Entity Alignment in Knowledge Graph Embedding},
  booktitle = {2020 {IEEE} Fifth International Conference on Data Science in Cyberspace},
  year = {2020},
  pages = {333--338}
}

@article{li2013gaussianPHD,
  author = {Wenling Li and Yingmin Jia and Junping Du and Fashan Yu},
  title = {Gaussian Mixture {PHD} Filter for Multi-Sensor Multi-Target Tracking with Registration Errors},
  journal = {Signal Processing},
  year = {2013},
  volume = {93},
  number = {1},
  pages = {86--99}
}

@inproceedings{yuan2020financialSentiment,
  author = {Xunpu Yuan and Yawen Li and Zhe Xue and Feifei Kou},
  title = {Financial Sentiment Analysis Based on Pre-Training and TextCNN},
  booktitle = {Chinese Intelligent Systems Conference},
  year = {2020},
  pages = {48--56}
}

@inproceedings{peng2020embedding,
  author = {Yuting Peng and Jian Zhang and Chuan Zhou and others},
  title = {Embedding-Based Entity Alignment Using Relation Structural Similarity},
  booktitle = {2020 {IEEE} International Conference on Knowledge Graph},
  year = {2020},
  pages = {123--130}
}

@article{tang2020bertInt,
  author = {Xiaotian Tang and Jing Zhang and Bo Chen and others},
  title = {{BERT-INT}: A {BERT}-Based Interaction Model for Knowledge Graph Alignment},
  journal = {Interactions},
  year = {2020},
  volume = {100},
  pages = {e1}
}

@article{yang2019aligning,
  author = {Hongwei Yang and Yang Zou and Peng Shi and others},
  title = {Aligning Cross-Lingual Entities with Multi-Aspect Information},
  journal = {arXiv preprint arXiv:1910.06575},
  year = {2019}
}

@article{guan2019selfLearning,
  author = {Shuang Guan and Xiaofei Jin and Yafang Wang and others},
  title = {Self-Learning and Embedding Based Entity Alignment},
  journal = {Knowledge and Information Systems},
  year = {2019},
  volume = {59},
  pages = {361--386}
}

@article{liu2021activeEA,
  author = {Bing Liu and H. Scells and Guido Zuccon and others},
  title = {{ActiveEA}: Active Learning for Neural Entity Alignment},
  journal = {arXiv preprint arXiv:2110.06474},
  year = {2021}
}

@inproceedings{mao2020relational,
  author = {Xin Mao and Wenting Wang and Huimin Xu and others},
  title = {Relational Reflection Entity Alignment},
  booktitle = {Proceedings of the 29th {ACM} International Conference on Information and Knowledge Management},
  year = {2020},
  pages = {1095--1104}
}

@inproceedings{li2020leveragingGraph,
  author = {Wei Li and Xinyan Xiao and Jiachen Liu and Hua Wu and Haifeng Wang and Junping Du},
  title = {Leveraging Graph to Improve Abstractive Multi-Document Summarization},
  booktitle = {Proceedings of the 58th Annual Meeting of the Association for Computational Linguistics},
  year = {2020},
  pages = {6232--6243}
}

@article{liang2020abstractiveSocial,
  author = {Zeyu Liang and Junping Du and Chaoyang Li},
  title = {Abstractive Social Media Text Summarization Using Selective Reinforced Seq2Seq Attention Model},
  journal = {Neurocomputing},
  year = {2020},
  volume = {410},
  pages = {432--440}
}

@article{meng2016consensus,
  author = {Deyuan Meng and Yingmin Jia and Junping Du},
  title = {Consensus Seeking via Iterative Learning for Multi-Agent Systems with Switching Topologies and Communication Time-Delays},
  journal = {International Journal of Robust and Nonlinear Control},
  year = {2016},
  volume = {26},
  number = {17},
  pages = {3772--3790}
}

@inproceedings{xu2016multiFeatureIndexing,
  author = {Zihang Xu and Junping Du and Lingfei Ye and Dan Fan},
  title = {Multi-Feature Indexing for Image Retrieval Based on Hypergraph},
  booktitle = {2016 4th International Conference on Cloud Computing and Intelligent Systems},
  year = {2016},
  pages = {494--500}
}

@article{yang2015ontology,
  author = {Yuehua Yang and Junping Du and Yuan Ping},
  title = {Ontology-Based Intelligent Information Retrieval System},
  journal = {Journal of Software},
  year = {2015},
  volume = {26},
  number = {7},
  pages = {1675--1687}
}

@article{xu2022topicTrend,
  author = {Mingying Xu and Junping Du and Zhe Xue and Zeli Guan and Feifei Kou and Lei Shi},
  title = {A Scientific Research Topic Trend Prediction Model Based on Multi-LSTM and Graph Convolutional Network},
  journal = {International Journal of Intelligent Systems},
  year = {2022}
}

@article{li2014phdGlint,
  author = {Wenling Li and Yingmin Jia and Junping Du and Jun Zhang},
  title = {{PHD} Filter for Multi-Target Tracking with Glint Noise},
  journal = {Signal Processing},
  year = {2014},
  volume = {94},
  pages = {48--56}
}

@article{xu2023infomax,
  author = {Xin Xu and Junping Du and Jie Song and Zhe Xue},
  title = {{InfoMax} Classification-Enhanced Learnable Network for Few-Shot Node Classification},
  journal = {Electronics},
  year = {2023},
  volume = {12},
  number = {1},
  pages = {239}
}

@inproceedings{chen2020stockAttention,
  author = {Jiannan Chen and Junping Du and Zhe Xue and Feifei Kou},
  title = {Prediction of Financial Big Data Stock Trends Based on Attention Mechanism},
  booktitle = {2020 {IEEE} International Conference on Knowledge Graph},
  year = {2020},
  pages = {152--156}
}

@article{cui2021mvgan,
  author = {Wanqiu Cui and Junping Du and Dawei Wang and Feifei Kou and Zhe Xue},
  title = {{MVGAN}: Multi-View Graph Attention Network for Social Event Detection},
  journal = {{ACM} Transactions on Intelligent Systems and Technology},
  year = {2021},
  volume = {12},
  number = {3},
  pages = {1--24}
}

@article{shao2021memoryAwareRandomWalk,
  author = {Yingxia Shao and Shiyue Huang and Yawen Li and Xupeng Miao and Bin Cui and Lei Chen},
  title = {Memory-Aware Framework for Fast and Scalable Second-Order Random Walk Over Billion-Edge Natural Graphs},
  journal = {The VLDB Journal},
  year = {2021},
  volume = {30},
  number = {5},
  pages = {769--797},
  language = {english}
}

@inproceedings{ou2024viewCategoryTransformer,
  author = {Shilong Ou and Zhe Xue and Yawen Li and Meiyu Liang and Yuanqiang Cai and Junjiang Wu},
  title = {View-Category Interactive Sharing Transformer for Incomplete Multi-View Multi-Label Learning},
  booktitle = {Proceedings of the {IEEE/CVF} Conference on Computer Vision and Pattern Recognition},
  year = {2024},
  pages = {27467--27476},
  language = {english}
}

@article{li2022stochasticQuantization,
  author = {Yawen Li and Wenling Li and Zhe Xue},
  title = {Federated Learning With Stochastic Quantization},
  journal = {International Journal of Intelligent Systems},
  year = {2022},
  volume = {37},
  number = {12},
  pages = {11600--11621},
  language = {english}
}

@article{zhou2020creativeEntrepreneur,
  author = {Jinyi Zhou and Xingzi Xu and Yawen Li and Chengcheng Liu},
  title = {Creative Enough to Become an Entrepreneur: A Multi-Wave Study of Creative Personality, Education, Entrepreneurial Identity, and Innovation},
  journal = {Sustainability},
  year = {2020},
  volume = {12},
  number = {10},
  pages = {4043},
  language = {english}
}

@inproceedings{zhang2025td3,
  author = {Jiaqing Zhang and Mingjia Yin and Hao Wang and Yawen Li and Yuyang Ye and Xingyu Lou and Junping Du and Enhong Chen},
  title = {{TD3}: Tucker Decomposition Based Dataset Distillation Method for Sequential Recommendation},
  booktitle = {Proceedings of the {ACM} on Web Conference 2025},
  year = {2025},
  pages = {3994--4003},
  language = {english}
}

@article{jin2022heterogeneousGraphContrastive,
  author = {Di Jin and Cuiying Huo and Jianwu Dang and Pengyang Zhu and Wenwu Zhang and Witold Pedrycz and Lingfei Wu},
  title = {Heterogeneous Graph Neural Networks Using Self-Supervised Reciprocally Contrastive Learning},
  journal = {arXiv preprint arXiv:2205.00256},
  year = {2022},
  language = {english}
}

\end{document}